\documentclass[letterpaper]{article}
\usepackage{aaai2026}

\usepackage[hyphens]{url}
\usepackage{listings}
\usepackage{graphicx}
\usepackage{natbib}
\usepackage{caption}
\usepackage{booktabs}
\usepackage{amsmath}
\usepackage{amssymb}
\usepackage{multirow}
\usepackage{subcaption}
\usepackage[colorlinks=true,linkcolor=blue,citecolor=blue,urlcolor=blue]{hyperref}
\usepackage{cleveref}
\usepackage{stfloats}

\nocopyright

\title{
\includegraphics[width=0.28\linewidth]{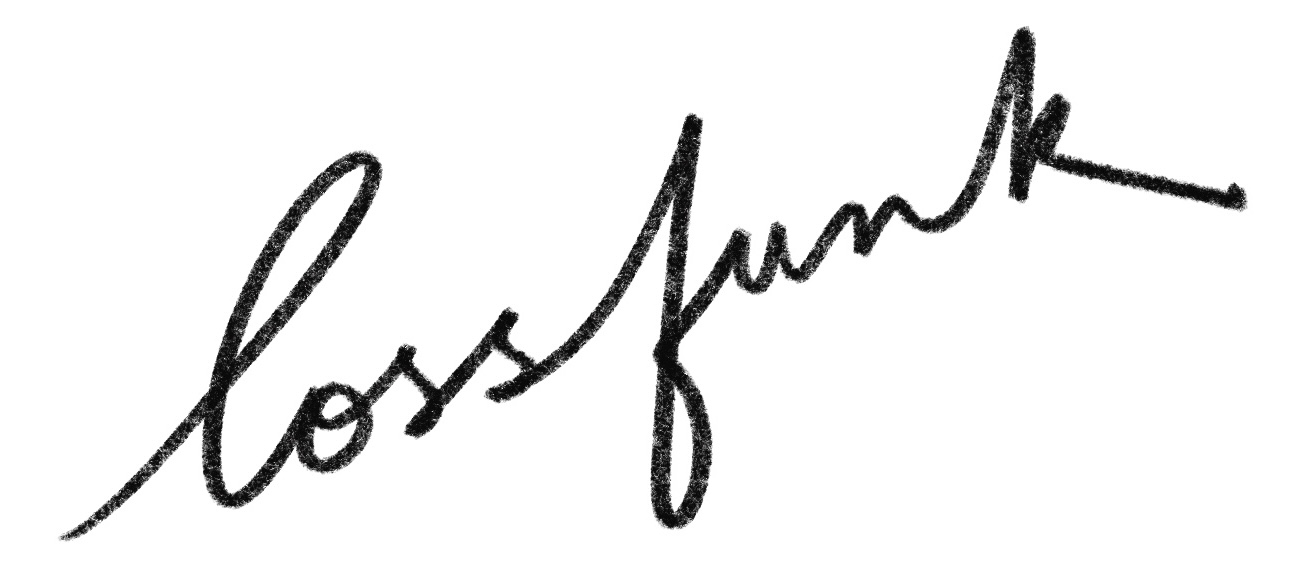}
\\[0.2cm]
See, Symbolize, Act: Grounding VLMs with Spatial Representations for Better Gameplay
}

\author{
Ashish Baghel, Paras Chopra
}

\affiliations{
Lossfunk\\
\{ashish.baghel, paras\}@lossfunk.com
}

\begin{document}
\raggedbottom
\maketitle
\vspace{-0.6em}

\begin{abstract}
Vision-Language Models (VLMs) excel at describing visual scenes, yet struggle to translate perception into precise, grounded actions. We investigate whether providing VLMs with both the visual frame and the symbolic representation of the scene can improve their performance in interactive environments. We evaluate three strong VLMs across Atari games, VizDoom, and AI2-THOR, comparing frame-only, frame with self-extracted symbols, frame with ground-truth symbols, and symbol-only pipelines.\textbf{ Our results indicate that all models benefit when the symbolic information is accurate}. However, when VLMs extract symbols themselves, performance becomes dependent on model capability and scene complexity. We further investigate how accurately VLMs can extract symbolic information from visual inputs and how noise in these symbols affects decision-making and gameplay performance.
Our findings reveal that symbolic grounding is beneficial in VLMs only when symbol extraction is reliable. Code and data are available at \url{https://github.com/Lossfunk/See-Symbolize-Act}.
\end{abstract}

\section{Introduction}

Vision-Language Models (VLMs) are increasingly used to build general-purpose AI agents that can both interpret visual scenes and decide how to act within them. When these systems shift from passive perception to interactive decision-making in fields like robotics, embodied AI, or game environments, they encounter tasks that rely on precise spatial understanding, an ability that current VLMs do not yet provide reliably \citep{waytowich2024atari}.

Atari gameplay provides a controlled setting for studying these spatial reasoning challenges. Games like Pong, Breakout, and Space Invaders require precise tracking of paddles, balls, and aliens. Under these conditions, current VLMs frequently misidentify objects, repeat ineffective actions, or fail to execute precise controls, lagging behind fine-tuned game-specific agents.

Prior work either fine-tunes VLMs on thousands of game-specific trajectories \citep{zhai2024finetuning}, sacrificing zero-shot generalization, or relies on visual frames alone, as in Atari-GPT \citep{waytowich2024atari}, where spatial reasoning remains a challenge. However, to the best of our knowledge, no systematic study has evaluated how accurately VLMs identify objects and coordinates, or how the coordinate accuracy affects the models' ability to take better actions as scene complexity increases.

To probe the capabilities of current VLMs, we conduct systematic evaluations across three strong VLMs (Claude-4-Sonnet, GPT-4o, Gemini-2.5-Pro) on Atari games with varying levels of complexity (Pong, Breakout, and Space Invaders). To isolate the contributing factors, we compare four pipelines: frame-only, frame with self-extracted symbols, frame with ground-truth symbols, and ground-truth symbol-only (no visual context).

In this paper, we analyze VLMs' ability to extract symbolic information from visual input and evaluate their accuracy across multiple games. We also investigate how the quality of extracted symbols influences decision-making, and determine when symbolic information benefits or harms agent performance.

\section{Related Work}

\paragraph{VLMs for Game Playing}
Recent work has explored the use of large Vision-Language Models as zero-shot agents in interactive environments. Atari-GPT \citep{waytowich2024atari} and TextAtari \citep{textatari2025} evaluate multimodal models on Atari games and show that, while VLMs can understand visual scenes, they struggle with consistent control and spatial precision during gameplay. Fine-tuning improves decision-making in VLM-based agents \citep{zhai2024finetuning}, but this requires task-specific data and does not resolve whether VLMs can act effectively in a purely zero-shot setting.

\paragraph{Object-Centric State Representations}
Object-centric state representations help simplify reasoning and planning in control tasks. OCAtari \citep{delfosse2023ocatari} extracts objects and their coordinates directly from Atari's RAM, providing ground-truth symbolic information from memory. Prior work has also shown that planning over symbolic or learned structured states benefits decision-making \citep{dittadi2021planning}. However, these approaches typically assume access to clean or near-perfect symbolic information. Our work differs by examining symbols extracted directly from vision in a zero-shot manner and studying how their quality impacts downstream actions.

\paragraph{Symbolic Grounding in Agents}
The symbolic grounding problem \citep{harnad1990} highlights the fundamental challenge of connecting high-level symbols to perception. ReAct \citep{yao2022react} demonstrated that combining reasoning and acting can improve tool-use-inspired language-driven action loops. More recent work focuses on improving spatial reasoning capabilities in VLMs \citep{chen2024spatialvlm,spatialrgpt2024}, and studies have analyzed why it remains a persistent challenge for such models \citep{whyspatial2025}. Robustness under noisy visual conditions has also been explored \citep{robustnessnoisy2025}, but prior work has not examined how perception errors propagate through a symbolic control loop. PoE-World \citep{piriyakulkij2025poeworld} combines language models with programmatic symbolic world models for Atari tasks, but assumes reliable symbolic input.

\section{Methodology}

Our experimental setup evaluates whether symbolic grounding helps VLMs' gameplay performance. We evaluate multiple Vision-Language Models across Atari games.

\subsection{Experimental Setup}

\textbf{Games.} We use Atari games that differ in object count and the spatial reasoning required: \textit{Pong}, \textit{Breakout}, and \textit{Space Invaders}. These games were chosen because their spatial control demands are known to challenge VLMs \citep{waytowich2024atari}. These games span different visual complexity levels: Pong (2--4 objects: ball, paddle), Breakout (5--15 objects: ball, paddle, bricks), and Space Invaders (20--50 objects: player, aliens, bullets, shields). These games allow us to test how symbolic grounding behaves as the gameplay becomes more complex.

\textbf{Models.} We evaluate VLMs: Claude-4-Sonnet, GPT-4o, and Gemini-2.5-Pro. All models were used in a zero-shot setting without any fine-tuning.

\subsection{Pipelines}

\begin{figure}[!htb]
    \centering
    \includegraphics[width=0.9\linewidth]{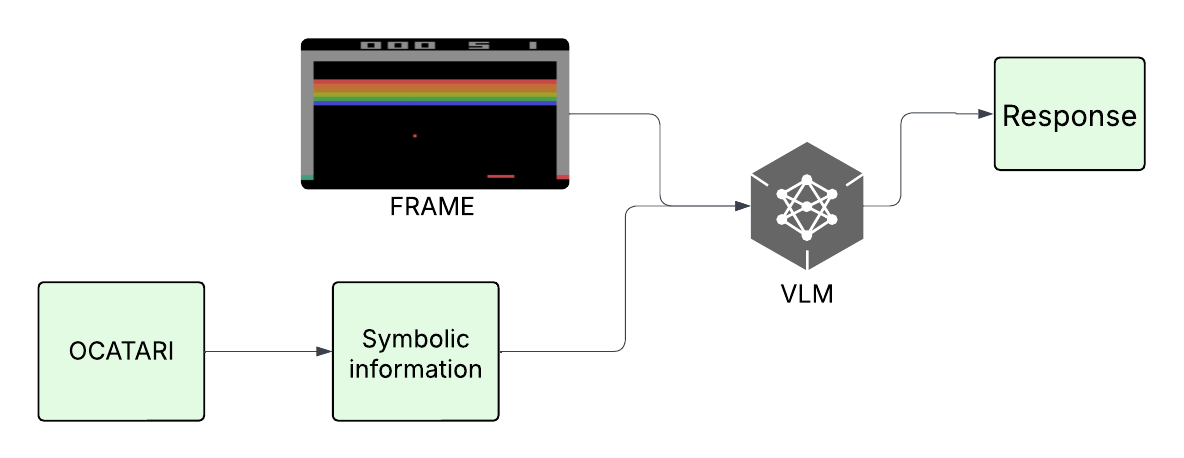}
    \caption{Frame + Ground-Truth Symbols pipeline.}
    \label{fig:ground_truth}
\end{figure}

\textbf{Frame + Ground-Truth Symbols (F+S-GT).} This pipeline reads symbolic information directly from the game RAM using OCAtari~\citep{delfosse2023ocatari}, which provides perfect object positions with zero detection error and serves as the upper bound for all other pipelines. The VLM uses both the frame and the symbols to select the next action, as illustrated in Figure~\ref{fig:ground_truth}.

\begin{figure}[!htb]
    \centering
    \includegraphics[width=0.9\linewidth]{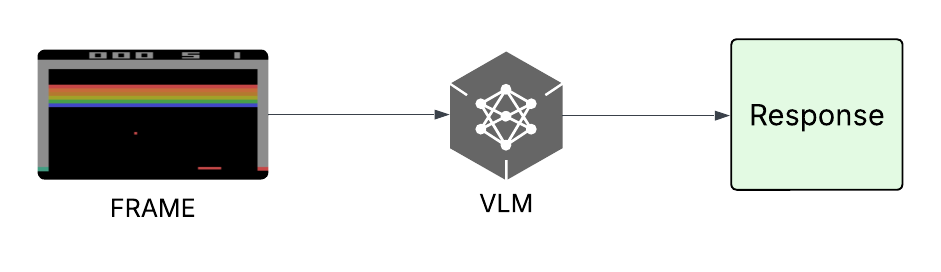}
    \caption{Frame-only pipeline.}
    \label{fig:vision_only}
\end{figure}

\textbf{Frame-only (F).} The VLM receives only the raw game frame and available action mappings, and outputs the action for the next frame. This setup tests whether the model can perform spatial reasoning using only visual input, without any symbolic help (Figure~\ref{fig:vision_only}).

\begin{figure}[!htb]
    \centering
    \includegraphics[width=0.9\linewidth]{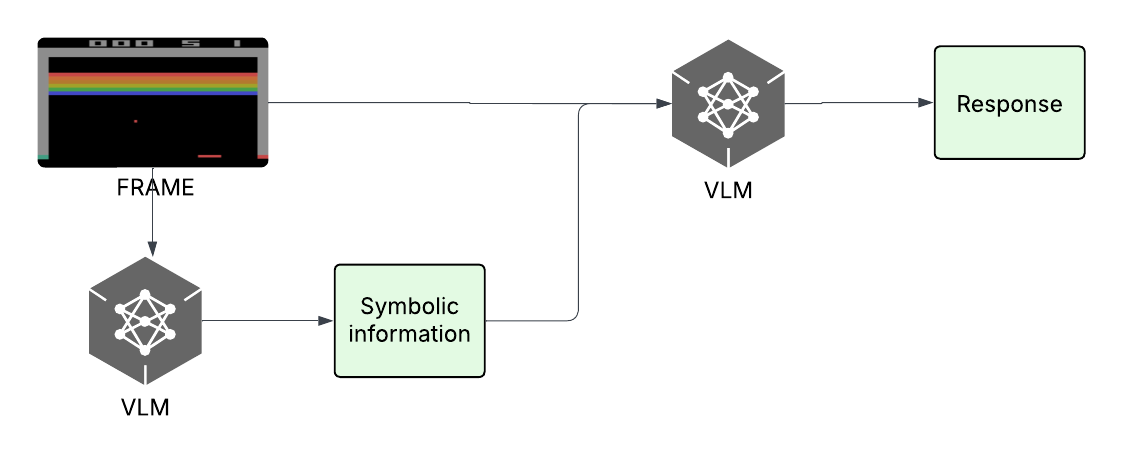}
    \caption{Frame + Self-Extracted Symbols pipeline.}
    \label{fig:vision_symbol}
\end{figure}

\textbf{Frame + Self-Extracted Symbols (F+S-self).} This pipeline follows a two-stage process, illustrated in Figure~\ref{fig:vision_symbol}. First, the VLM extracts object-centric symbolic information from the frame as structured data, including object IDs, labels, $(x, y)$ coordinates, and confidence scores. The VLM then uses both the frame and the extracted symbols to select the next action. The detailed steps of the symbol extraction process and its limitations are discussed later in the Symbolic Detection Quality subsection, where we analyze \textit{why} symbolic grounding helps or hurts performance.

\begin{figure}[!htb]
    \centering
    \includegraphics[width=0.9\linewidth]{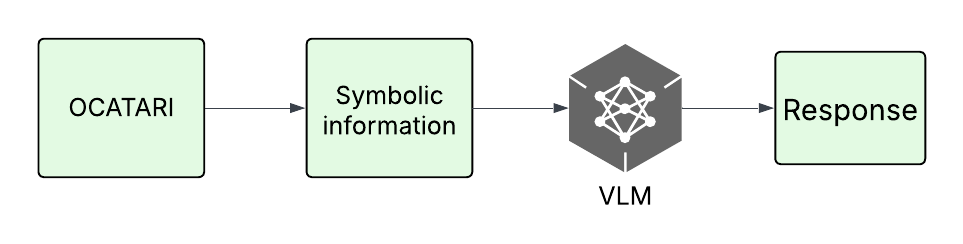}
    \caption{Symbol-only pipeline.}
    \label{fig:symbol_only}
\end{figure}

\textbf{Symbol-Only (S-GT).} The VLM receives object coordinates from the RAM state, with \textit{no visual frame}, as shown in Figure~\ref{fig:symbol_only}. This pipeline isolates the contribution of symbolic information and tests whether VLMs can play Atari by reasoning purely with symbols that provide perfect object detection and no visual context at all. This reveals whether symbols alone are sufficient or whether visual grounding is essential for VLM decision-making.

\subsection{Evaluation Metrics}
\label{sec:detection}
\textbf{Gameplay Metric.} We use cumulative reward as the gameplay metric. The game is paused at every frame, and the VLM decides the next action. Each model plays 600 frames per run, and uses a frame size of $1280 \times 720$, across 2 seeds. Atari frames are upscaled from the native resolution ($160 \times 210$) to $1280 \times 720$ using bilinear interpolation. The same interpolation method is applied across all resolution conditions in the ablation study. VizDoom and AI2-THOR experiments use the same seed count. This setup isolates reasoning quality, since real-time latency is not the focus of this study. To compare models on a common scale, we normalize raw scores relative to each model's Frame + GT Symbols performance. The lower bound is the worst observed: -17 for Pong, $0$ for Breakout, and $0$ for Space Invaders over 600 frames. Absolute scores are reported in the results tables.

\textbf{Detection Metrics.} To better understand \textit{why} different pipelines succeed or fail, we measure symbolic extraction quality by comparing VLM-generated symbols with ground-truth annotations across 100 frames per game. These frames come from a dedicated evaluation set designed to include hard cases (e.g., ball near corners, dense bricks, aliens at screen edges) that gameplay rollouts may not reliably reach, so the reported metrics reflect a stress test of detection rather than typical-case accuracy. We use standard detection metrics: (1) \textbf{F1 score}, which gives a balanced measure of how many objects are correctly identified, penalizing both missed detections and false positives; and (2) \textbf{IoU (overlap score)}, which measures how closely the predicted object coordinates align with the ground-truth positions.

\subsubsection{Prompt Structure}
We use a generic prompt template for every pipeline, supplying only the actions and the raw inputs (frame, symbols, or both). This ensures that the model receives no task-specific instructions, examples, or strategy hints. The exact wording for the Frame + Self-Extracted Symbols condition is given below; the other variants differ only by the presence or absence of the image or symbol block. Complete prompts for all conditions are listed in Appendix~\ref{sec:appendix_prompts}.

\begin{table}[!htb]
\centering
\caption{Example prompt structure for Frame + Self-Extracted Symbols pipeline (abbreviated).}
\label{tab:prompt_example}
\small
\begin{tabular}{p{7.5cm}}
\toprule
\textbf{Frame + Self-Extracted Symbols Prompt} \\
\midrule
\texttt{You are an expert [game] player analyzing a game frame.} \\
\texttt{} \\
\texttt{Game controls:} \\
\texttt{- Action 0: NOOP, Action 1: FIRE, ...} \\
\texttt{} \\
\texttt{Current frame analysis:} \\
\texttt{- Total objects detected: 7} \\
\texttt{} \\
\texttt{Detected objects with coordinates:} \\
\texttt{- Object 'ball': x=580, y=450, size 15x20} \\
\texttt{- Object 'paddle': x=850, y=650, size 135x15} \\
\texttt{- Object 'brick\_wall': x=640, y=260, size 1140x120} \\
\texttt{- ...} \\
\texttt{} \\
\texttt{IMPORTANT: Use symbolic information when reliable,} \\
\texttt{but prioritize visual reasoning if data is incomplete.} \\
\texttt{} \\
\texttt{Choose the optimal action. Return JSON:} \\
\texttt{\{"reasoning": "...", "action": integer\}} \\
\bottomrule
\end{tabular}
\end{table}

\section{Results}

\subsection{Main Finding: Model-Specific Effects of Symbolic Grounding}

We find that symbolic grounding helps only when a model can extract accurate symbols, and this ability breaks down differently across models and game complexity. Table~\ref{tab:main_results} shows this pattern clearly.

The improvement is most pronounced for Claude-4-Sonnet. In the frame-only pipeline, Claude performs poorly in all games. It scores $-16.0$ in \textit{Pong} and $0.0$ in \textit{Breakout}, repeatedly taking the same incorrect actions because it misidentifies the paddle and its position in both games. When we provide self-extracted symbols, Claude's performance improves substantially: It reaches $-3.0$ in \textit{Pong} and $12.0$ in \textit{Breakout}, essentially matching the upper bound. Even in the visually dense \textit{Space Invaders} environment, symbols increase Claude's score from $80.0$ to $150.0$, as the model is now able to take more informed and effective actions.

GPT-4o and Gemini-2.5-Pro behave differently. In the frame-only pipeline, both achieve stronger baseline scores than Claude, indicating better action selection from visual input alone. Adding symbols results in only modest gains in the simpler games: Gemini improves to $-3.0$ in \textit{Pong} and $10.0$ in \textit{Breakout}, while GPT-4o reaches $-6.5$ in Pong and $8.0$ in Breakout. In \textit{Space Invaders}, however, both models show substantial performance degradation. The 20--50 objects on screen overwhelm the detection capabilities of GPT-4o and Gemini, whose lower extraction accuracy (F1 of 0.124 and 0.189, respectively; see Table~\ref{tab:detection_metrics}) produces many incorrect coordinates that mislead their reasoning and negate potential gains.

\begin{table}[!htb]
\centering
\caption{
Average reward over 600 frames across pipelines.
Abbreviations:
F = Frame-only;
S-GT = Symbol-only;
F+S-self = Frame + Self-Extracted Symbols;
F+S-GT = Frame + Ground-Truth Symbols.
}
\label{tab:main_results}
\footnotesize
\setlength{\tabcolsep}{3pt}
\begin{tabular}{llccc}
\toprule
\textbf{Model} & \textbf{Pipeline} & \textbf{Pong} & \textbf{Breakout} & \textbf{Space Inv.} \\
\midrule

\multirow{4}{*}{Claude-4-Sonnet}
    & F+S-GT & $-1.0$ & $12.0$ & $175.0$ \\
    & F & $-16.0$ & $0.0$ & $80.0$ \\
    & S-GT & $-14.0$ & $0.0$ & $90.0$ \\
    & F+S-self & $-3.0$ & $12.0$ & $150.0$ \\

\midrule

\multirow{4}{*}{Gemini-2.5-Pro}
    & F+S-GT & $-1.0$ & $12.0$ & $170.0$ \\
    & F & $-7.0$ & $7.0$ & $95.0$ \\
    & S-GT & $-12.0$ & $3.0$ & $95.0$ \\
    & F+S-self & $-3.0$ & $10.0$ & $80.0$ \\

\midrule

\multirow{4}{*}{GPT-4o}
    & F+S-GT & $-3.0$ & $13.0$ & $185.0$ \\
    & F & $-5.0$ & $7.5$ & $130.0$ \\
    & S-GT & $-14.0$ & $0.0$ & $105.0$ \\
    & F+S-self & $-6.5$ & $8.0$ & $65.0$ \\

\bottomrule
\end{tabular}
\end{table}

The Symbol-only pipeline shows that visual grounding is essential. Even when models receive perfect symbolic coordinates identical to ground truth, removing the visual frame causes performance to collapse. In Breakout, Claude-4-Sonnet and GPT-4o both score zero points. In Space Invaders, in the best case, GPT-4o drops from 185 points (with Frame + GT Symbols) to just 105 (57\% of the upper bound) without visual context. This proves that symbolic information alone is insufficient. Visual frames act as essential scaffolding, giving VLMs the perceptual context they need to correctly interpret and trust coordinate data.

\begin{figure}[!htb]
\centering
\includegraphics[width=\linewidth]{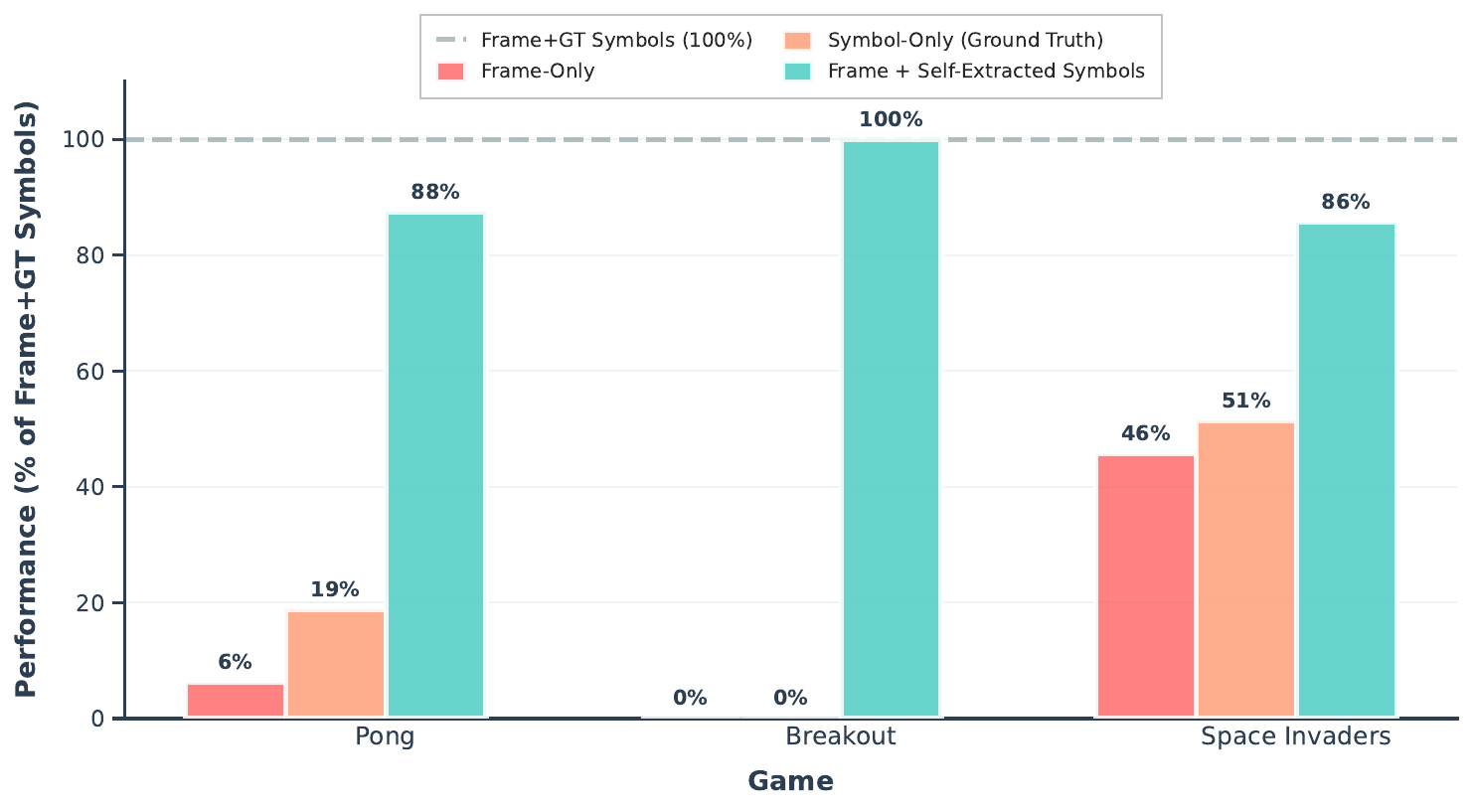}
\caption{Claude-4-Sonnet performance across games, normalized relative to each model's F+S-GT score (see Table~\ref{tab:main_results} for absolute values). Claude shows consistent improvement with self-extracted symbols across all games, achieving near-optimal performance in Breakout.}
\label{fig:model_performance_individual}
\end{figure}

The Frame + Ground-Truth Symbols results show that when models receive perfect symbols derived from the RAM state and visual frame, their performance consistently improves, confirming that accurate symbolic information is always beneficial. To evaluate how close each model gets to its upper bound with symbolic grounding, Figure~\ref{fig:model_performance_individual} shows detailed model-wise performance relative to the Frame + Ground-Truth Symbols.

\subsection{Symbolic Detection Quality Across VLMs}
\label{sec:symbol_quality}

We find that the effect of object-centric information on gameplay depends directly on object detection accuracy. Table~\ref{tab:detection_metrics} shows that Claude-4-Sonnet achieves substantially higher extraction accuracy than Gemini-2.5-Pro and GPT-4o, which explains why symbolic grounding helps some models but hurts others.

\begin{table}[!htb]
\centering
\caption{Object detection metrics across models, averaged over 100 curated evaluation frames per game (see Section~\ref{sec:detection}).}
\label{tab:detection_metrics}
\begin{tabular}{lcc}
\toprule
\textbf{Model} & \textbf{F1 Score} & \textbf{IoU}  \\
\midrule
Claude-4-Sonnet & 0.715 & 0.533  \\
Gemini-2.5-Pro & 0.189 & 0.202  \\
GPT-4o & 0.124 & 0.128  \\
\bottomrule
\end{tabular}
\end{table}

Claude-4-Sonnet achieves an F1 score of 0.715 and an IoU of 0.533, indicating relatively accurate object detection and localization. This accuracy enables the Frame + Self-Extracted Symbols pipeline to provide useful information that improves model performance. In contrast, Gemini-2.5-Pro and GPT-4o achieve much lower F1 scores (0.189 and 0.124), missing and mislocating the majority of objects. These errors reduce the reliability of the symbolic representations and degrade decision-making performance, as shown in their model-wise breakdowns (Figure~\ref{fig:gemini_performance} and Figure~\ref{fig:gpt4o_performance}).

\subsection{Beyond Atari: Complex Visual Environments}
\label{sec:beyond_atari}

To evaluate whether our findings extend beyond simple 2D environments, we test the same pipelines on two more visually complex domains: VizDoom (a first-person shooter) and AI2-THOR (an embodied AI kitchen task).

\textbf{VizDoom.} We use the \textit{defend\_the\_center} scenario, where the agent must shoot approaching enemies from a fixed position. This environment features 3D graphics, textured surfaces, and variable enemy appearances, presenting substantially different visual challenges than Atari's pixel-art style.

\textbf{AI2-THOR.} We use a kitchen scene where the agent must collect food items and place them on a designated surface. This photorealistic environment requires understanding complex object relationships and 3D spatial reasoning.

For VizDoom, ground-truth symbols are obtained through the engine's labels buffer, which provides object identities, bounding boxes, and coordinates. For AI2-THOR, symbols are extracted from the simulator's object metadata API, which exposes each object's type, 3D position, and bounding box coordinates. In both cases, the symbolic information is derived directly from the engine's internal state.

\begin{table}[!htb]
\centering
\caption{VizDoom \textit{defend\_the\_center} performance (kills over 600 frames). Higher is better.}
\label{tab:vizdoom_results}
\small
\setlength{\tabcolsep}{4pt}
\begin{tabular}{lcccc}
\toprule
\textbf{Model} & \textbf{F} & \textbf{F+S-self} & \textbf{F+S-GT} & \textbf{S-GT} \\
\midrule
Claude-4-Sonnet & 5 & 9 & \textbf{14} & 12 \\
GPT-4o & 12 & 8 & \textbf{13} & 3 \\
Gemini-2.5-Pro & 11 & 4 & \textbf{13} & 12 \\
\bottomrule
\end{tabular}
\end{table}

\begin{table}[!htb]
\centering
\caption{AI2-THOR kitchen task performance (cumulative reward over 300 frames). Higher is better.}
\label{tab:ai2thor_results}
\small
\setlength{\tabcolsep}{4pt}
\begin{tabular}{lcccc}
\toprule
\textbf{Model} & \textbf{F} & \textbf{F+S-self} & \textbf{F+S-GT} & \textbf{S-GT} \\
\midrule
Claude-4-Sonnet & $-1.0$ & $2.0$ & $\mathbf{11.0}$ & $3.0$ \\
GPT-4o & $7.0$ & $9.0$ & $\mathbf{9.0}$ & $-1.0$ \\
Gemini-2.5-Pro & $5.0$ & $1.0$ & $\mathbf{10.0}$ & $-3.0$ \\
\bottomrule
\end{tabular}
\end{table}

\begin{figure}[!htb]
\centering
\includegraphics[width=\linewidth]{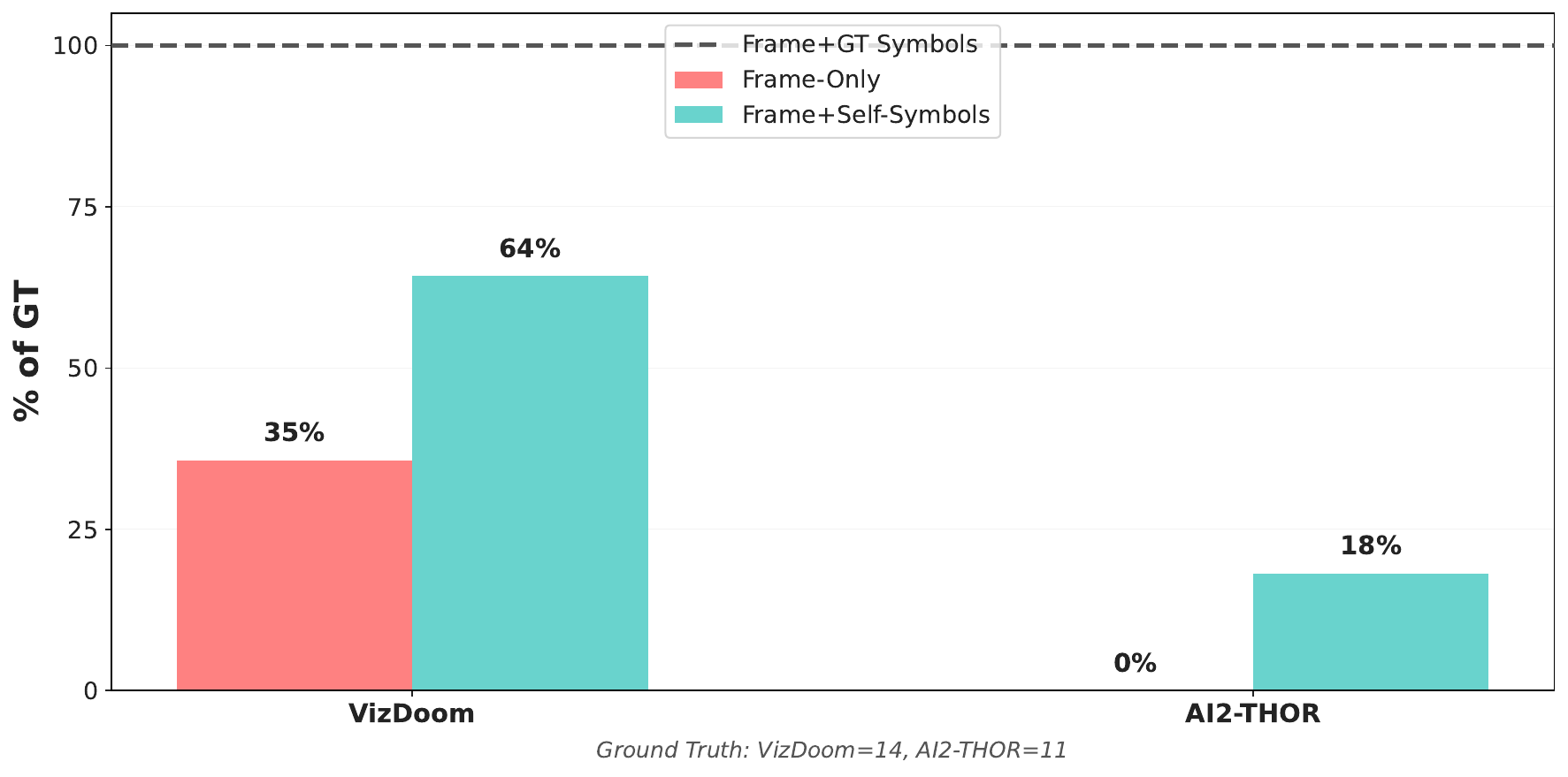}
\caption{Claude-4-Sonnet performance on complex 3D environments, normalized relative to F+S-GT score (see Tables~\ref{tab:vizdoom_results} and~\ref{tab:ai2thor_results} for absolute values). Claude improves significantly in VizDoom (35\%$\rightarrow$64\%) with self-extracted symbols.}
\label{fig:3d_environments_claude}
\end{figure}

\begin{figure}[!htb]
\centering
\includegraphics[width=\linewidth]{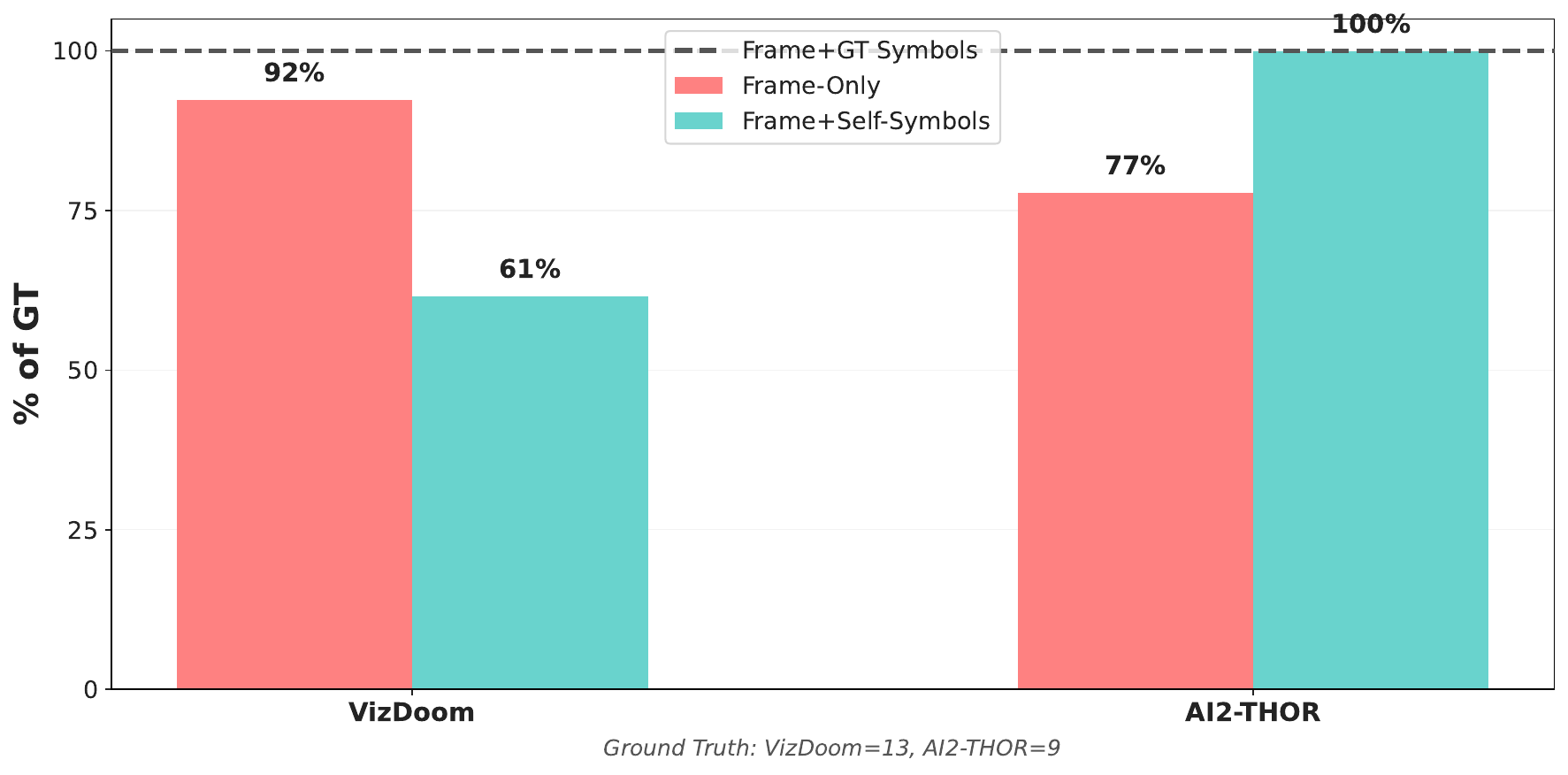}
\caption{GPT-4o performance on complex 3D environments, normalized relative to F+S-GT score (see Tables~\ref{tab:vizdoom_results} and~\ref{tab:ai2thor_results} for absolute values). GPT-4o matches the F+S-GT baseline in AI2-THOR with self-extracted symbols, demonstrating that accurate symbol extraction enables near-optimal performance.}
\label{fig:3d_environments_gpt4o}
\end{figure}

The results in Table ~\ref{tab:vizdoom_results}, ~\ref{tab:ai2thor_results} and Figures ~\ref{fig:3d_environments_claude} and ~\ref{fig:3d_environments_gpt4o} confirm our core finding: Frame + Ground-Truth Symbols (F+S-GT) consistently achieves the best performance across all models in both environments. In VizDoom, providing accurate symbolic information about enemy positions improves kill counts by 1--9 kills compared to frame-only baselines. In AI2-THOR, the improvement is even more pronounced, with F+S-GT scores reaching 9--11 compared to frame-only scores of $-1$ to $7$.

The Frame + Self-Extracted Symbols pipeline shows mixed results, consistent with our Atari findings. In VizDoom, Claude improves from 5 to 9 kills with self-extracted symbols, while GPT-4o and Gemini degrade (from 12 to 8 and 11 to 4, respectively). In AI2-THOR, Claude shows improvement from $-1$ to $2$ with self-extracted symbols, and GPT-4o achieves near-GT performance ($9.0$). This pattern confirms that symbolic grounding helps when models can extract accurate symbols from the environment.

The Symbol-only pipeline shows that visual context generally improves performance, though the effect is not uniform. Performance drops substantially without visual frames for GPT-4o in both environments. However, in VizDoom and AI2-THOR, symbol-only performance exceeds that of Frame + Self-Extracted Symbols for some models (e.g., Claude and Gemini in VizDoom), suggesting that when self-extracted symbols are inaccurate, they can be more harmful than having no visual frame at all. This indicates that the value of visual context depends on the quality of the symbolic information it is paired with.

\section{Ablation Studies}

We conduct two ablation studies to examine when symbolic information improves gameplay performance.

\subsection{How Input Resolution Affects Object Detection Accuracy}

Having established that detection quality determines whether symbolic information improves or degrades gameplay performance, we next examine how image resolution influences VLMs' ability to extract accurate object coordinates.

\subsubsection{Methodology}

We evaluate Claude-4-Sonnet's object detection and localization quality across four input resolutions on 100 frames per game:

\begin{itemize}
\item \textbf{Original ($160 \times 210$)}: Native Atari resolution with no scaling.
\item \textbf{Uniform $4\times$ ($640 \times 840$)}: Uniformly scaled while preserving the original aspect ratio.
\item \textbf{Square ($720 \times 720$)}: Square frame that distorts the original 4:3 aspect ratio.
\item \textbf{Current ($1280 \times 720$)}: The resolution used in our experiments, scaled by $8\times$ in width and $4\times$ in height.
\end{itemize}

Object extraction quality is measured using F1 score and IoU, and then compared against OCAtari ground-truth annotations.

\subsubsection{Results}

\begin{table}[!htb]
\centering
\caption{Detection quality across input resolutions for Claude-4-Sonnet (averaged 100 frames per game).}
\label{tab:resolution}
\footnotesize
\setlength{\tabcolsep}{3pt}
\begin{tabular}{lcccc}
\toprule
\textbf{Game} & \textbf{160$\times$210} & \textbf{640$\times$840} & \textbf{720$\times$720} & \textbf{1280$\times$720} \\
\midrule
\multicolumn{5}{c}{\textbf{F1 Score (higher indicates better performance)}} \\
\midrule
Pong & 0.28 & 0.52 & 0.55 & \textbf{0.58} \\
Breakout & 0.31 & 0.68 & 0.70 & \textbf{0.71} \\
Space Inv. & 0.35 & 0.71 & 0.73 & \textbf{0.75} \\
\midrule
\textbf{Average} & 0.31 & 0.64 & 0.66 & \textbf{0.68} \\
\bottomrule
\end{tabular}
\end{table}

\begin{figure}[!htb]
\centering
\includegraphics[width=\linewidth]{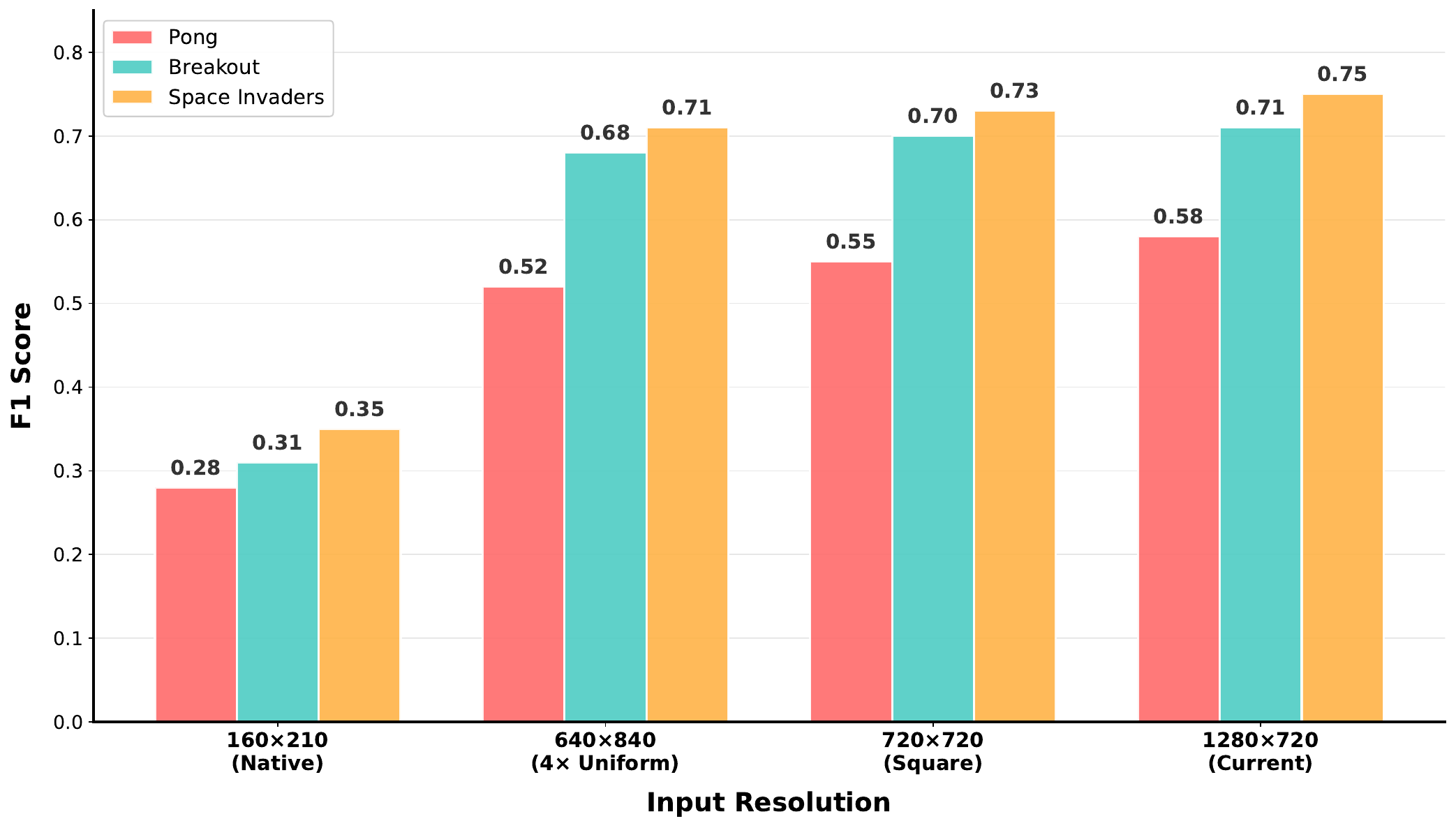}
\caption{Detection quality (F1 score) across resolutions and games. The native resolution ($160 \times 210$) produces lower detection accuracy (F1 $<$ 0.35), while higher resolutions enable effective symbolic grounding (F1 $>$ 0.65).}
\label{fig:resolution_f1}
\end{figure}

The native Atari resolution has low detection accuracy (average F1: 0.31). At this level, extracted symbols are too unreliable to improve gameplay and instead harm performance. Increasing resolution improves extraction quality, with $1280 \times 720$ achieving an average F1 of 0.68, approximately twice the original value. We observe that results beyond 720px on the shorter side: $720 \times 720$ (0.66) and $1280 \times 720$ (0.68) perform similarly. This suggests that total pixel count and detail preservation may be more important than preserving the original aspect ratio, as shown in Figure~\ref{fig:resolution_f1}.

\subsection{Noise Ablation: Robustness to Detection Errors}

We study how different levels of noise in object coordinates affect the usefulness of Object-centric information. This isolates the sensitivity of action selection to detection errors.

\subsubsection{Experimental Design}

We focus on Claude-4-Sonnet and GPT-4o, the two models that differ substantially in object detection quality.

The experiment injects Gaussian noise into ground-truth coordinates to simulate different levels of detection error.

\begin{equation}
x' = x + \mathcal{N}(0, \sigma \times W), \quad y' = y + \mathcal{N}(0, \sigma \times H)
\end{equation}
Here, $W$ and $H$ denote the frame width and height, respectively, where $\sigma$ controls noise magnitude. The five noise levels tested are:

\begin{itemize}
\item $\sigma = 0.00$: no-noise baseline
\item $\sigma = 0.10$: Low noise ($\sim$16--20 pixels of coordinate error)
\item $\sigma = 0.20$: Moderate noise ($\sim$32--40px)
\item $\sigma = 0.30$: High noise ($\sim$48--60px)
\item $\sigma = 0.40$: Severe noise ($\sim$64--80px)
\end{itemize}

Noise is sampled independently per frame and per object, replicating the randomness present in real-world visual frames.

For each model, game, and noise level, we run 10 independent seeds, each covering 300 frames. This reduces the variance and provides reliable mean performance estimates with confidence intervals.

\subsubsection{Results}

\begin{table}[t]
\centering
\caption{Performance under coordinate noise (mean reward across 10 seeds).}
\label{tab:noise_ablation}
\footnotesize
\setlength{\tabcolsep}{3pt}
\resizebox{\columnwidth}{!}{
\begin{tabular}{l l c c c c c}
\toprule
Game & Model & $\sigma$=0.0 & $\sigma$=0.1 & $\sigma$=0.2 & $\sigma$=0.3 & $\sigma$=0.4 \\
\midrule
Pong & Claude-4-Sonnet & -0.6 & -2.2 & -2.2 & -2.2 & -2.2 \\
Pong & GPT-4o & -2.7 & -3.0 & -3.4 & -3.8 & -2.8 \\
\midrule
Breakout & Claude-4-Sonnet & 5.0 & 4.3 & 3.4 & 3.4 & 2.8 \\
Breakout & GPT-4o & 5.0 & 4.0 & 3.0 & 2.3 & 2.6 \\
\midrule
Space Inv. & Claude-4-Sonnet & 53 & 57 & 57 & 88 & 52 \\
Space Inv. & GPT-4o & 86 & 43 & 69 & 66 & 62 \\
\bottomrule
\end{tabular}
}
\end{table}

\begin{figure}[!htb]
\centering
\includegraphics[width=0.9\linewidth]{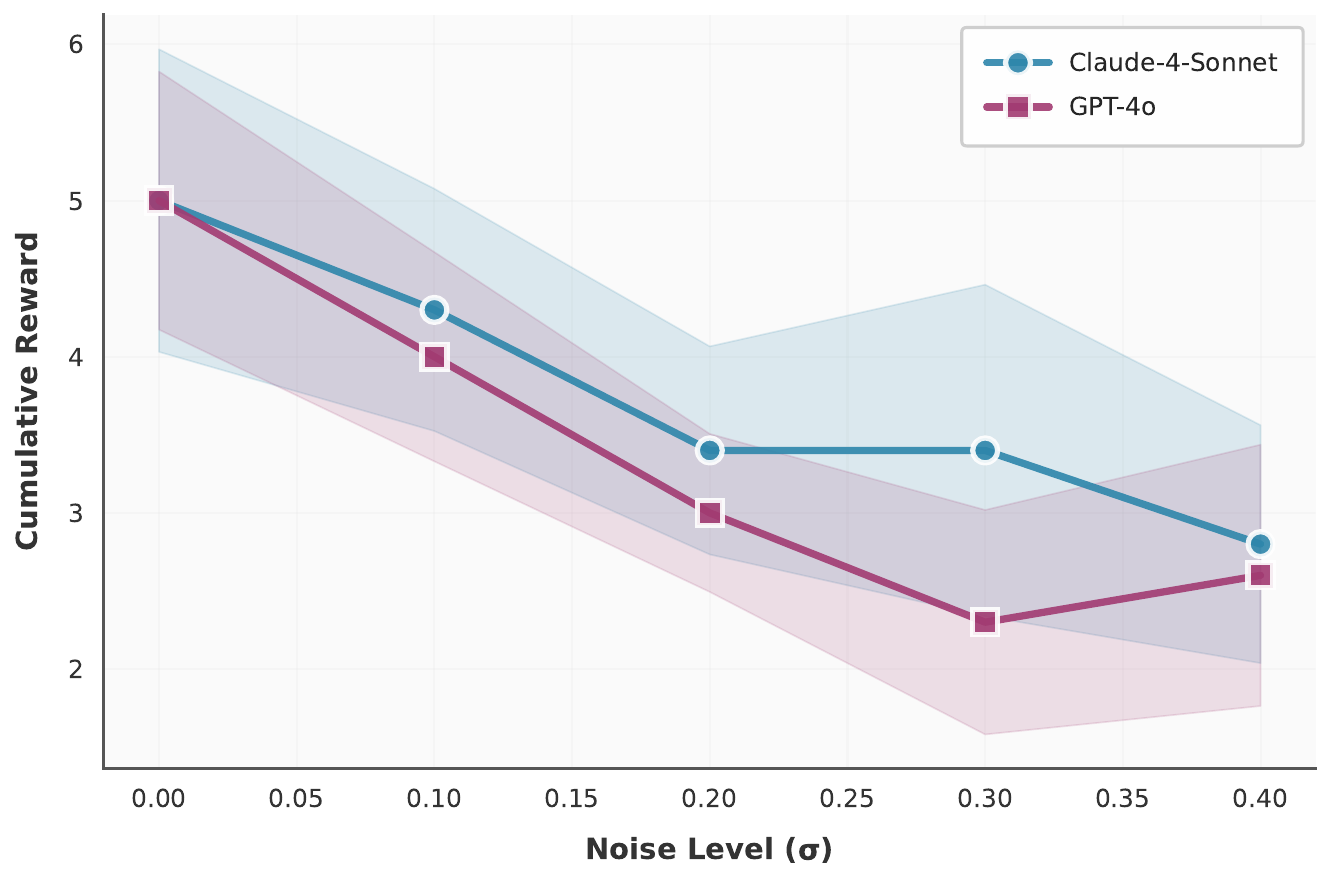}
\caption{Noise robustness analysis for Breakout. Both Claude-4-Sonnet and GPT-4o degrade significantly with even low noise ($\sigma=0.1$). Bars mark 95\% confidence intervals over 10 seeds.}
\label{fig:noise_by_game}
\end{figure}

Figure~\ref{fig:noise_by_game} illustrates the impact of coordinate noise on Breakout performance. In games like Pong and Breakout, both models show similar degradation: even a low level of noise causes a significant drop in performance. At $\sigma=0.1$ (just 16--20 pixels of error), scores decline by roughly 30--40\% relative to the no-noise baseline. Claude drops from $-0.6$ to $-2.2$ in Pong while GPT-4o decreases from $-2.7$ to $-3.0$ (Figure~\ref{fig:noise_pong}), and a similar decline happens in Breakout. By $\sigma=0.2$, the symbolic pipeline offers no clear advantage over using the frame alone.

Space Invaders behaves differently (Figure~\ref{fig:noise_spaceinvaders}). The dense formation of 20--50 aliens means even random firing can hit targets and score points, which partly explains the high variance in results. Claude's scores vary from $53$ to $88$ points across noise levels, whereas GPT-4o decreases sharply from $86$ (no noise) to $43$ at $\sigma=0.1$, losing half its performance from minimal noise.

\section{Limitations and Future Work}

While our findings demonstrate when symbolic grounding helps or harms VLMs, several limitations remain that open avenues for future work.

\textbf{Limited Environment Diversity.} While we extend our evaluation beyond Atari to include VizDoom and AI2-THOR, demonstrating that our findings hold in 3D and photorealistic environments, these represent only a subset of possible domains. Future work should explore additional environments with different visual and control characteristics.

\textbf{Limited Seed Count.} Our main results (Table~\ref{tab:main_results}) use two seeds per configuration, which limits statistical power. Given the inherent variance in Atari gameplay, small performance differences should be interpreted with caution. Future work should increase the number of seeds and report confidence intervals to strengthen the reliability of the main comparisons.

\textbf{Symbol Quality Bottleneck.} Our results show that the effectiveness of symbolic grounding is highly sensitive to symbol extraction quality. Future work should explore more robust symbol-extraction methods, such as hybrid detectors or lightweight fine-tuning of vision modules, to improve object-centric representations.

\textbf{Stateless Decision-Making.} Our setup treats each frame independently: the VLM receives only the current frame and symbols, with no access to previous frames or action history.

\textbf{Cost and Practicality.} Querying VLMs per frame incurs substantial latency and cost, making real-time gameplay infeasible with current models. Scaling symbolic grounding to real-time agents will therefore require fundamental efficiency improvements.

\section{Conclusion}
In this work, we set out to understand when symbolic grounding provides measurable benefits to VLMs in interactive environments. Our results, consistent across Atari games, VizDoom, and AI2-THOR, show that symbolic information is useful only when the extracted symbols are sufficiently accurate. When object positions are reliable, they provide the model with spatial information that frame-only reasoning struggles to capture. However, when symbols are noisy, they introduce ambiguity into the model's reasoning and lead to degraded decision-making. Even perfect symbolic information is insufficient without visual context. Future work should explore more robust symbol-extraction methods to unlock the full potential of symbolic grounding.

\bibliography{references}

\newpage
\appendix
\section{Appendix}
\label{sec:appendix_prompts}

\subsection{Additional Model Performance Results} The following figures show the complete model-wise performance breakdown for Gemini-2.5-Pro and GPT-4o, complementing the Claude-4-Sonnet results in the main paper.

\begin{figure}[!htb]
\centering
\includegraphics[width=0.9\linewidth]{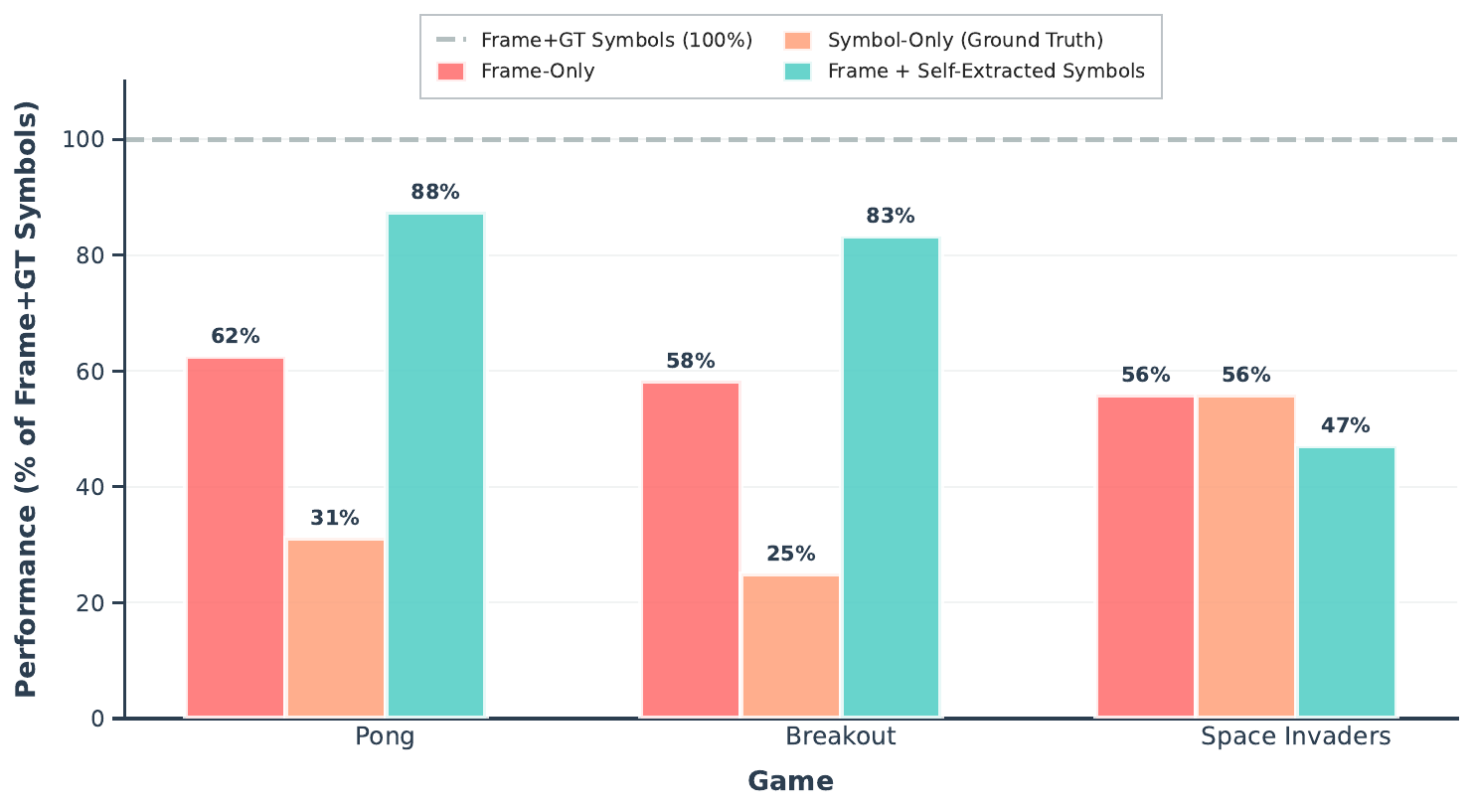}
\caption{Gemini-2.5-Pro performance breakdown, normalized relative to F+S-GT score (see Table~\ref{tab:main_results} for absolute values). Gemini benefits in simpler games but struggles in Space Invaders.}
\label{fig:gemini_performance}
\end{figure}

\begin{figure}[!htb]
\centering
\includegraphics[width=0.9\linewidth]{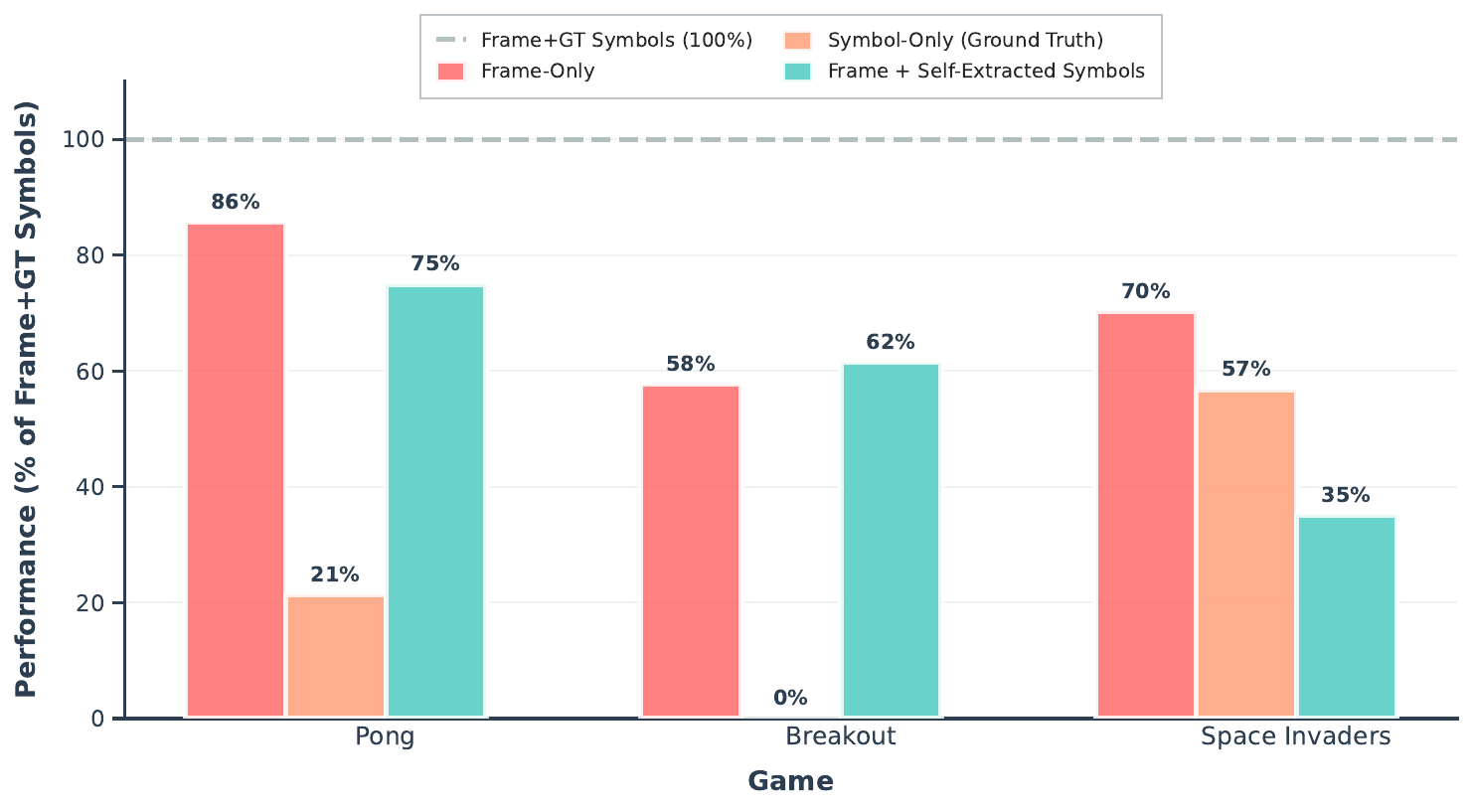}
\caption{GPT-4o performance breakdown, normalized relative to F+S-GT score (see Table~\ref{tab:main_results} for absolute values). GPT-4o shows mixed results, with symbolic information sometimes harming performance.}
\label{fig:gpt4o_performance}
\end{figure}

\subsection{Additional Noise Robustness Results}
\begin{figure}[!htb]
\centering
\includegraphics[width=0.9\linewidth]{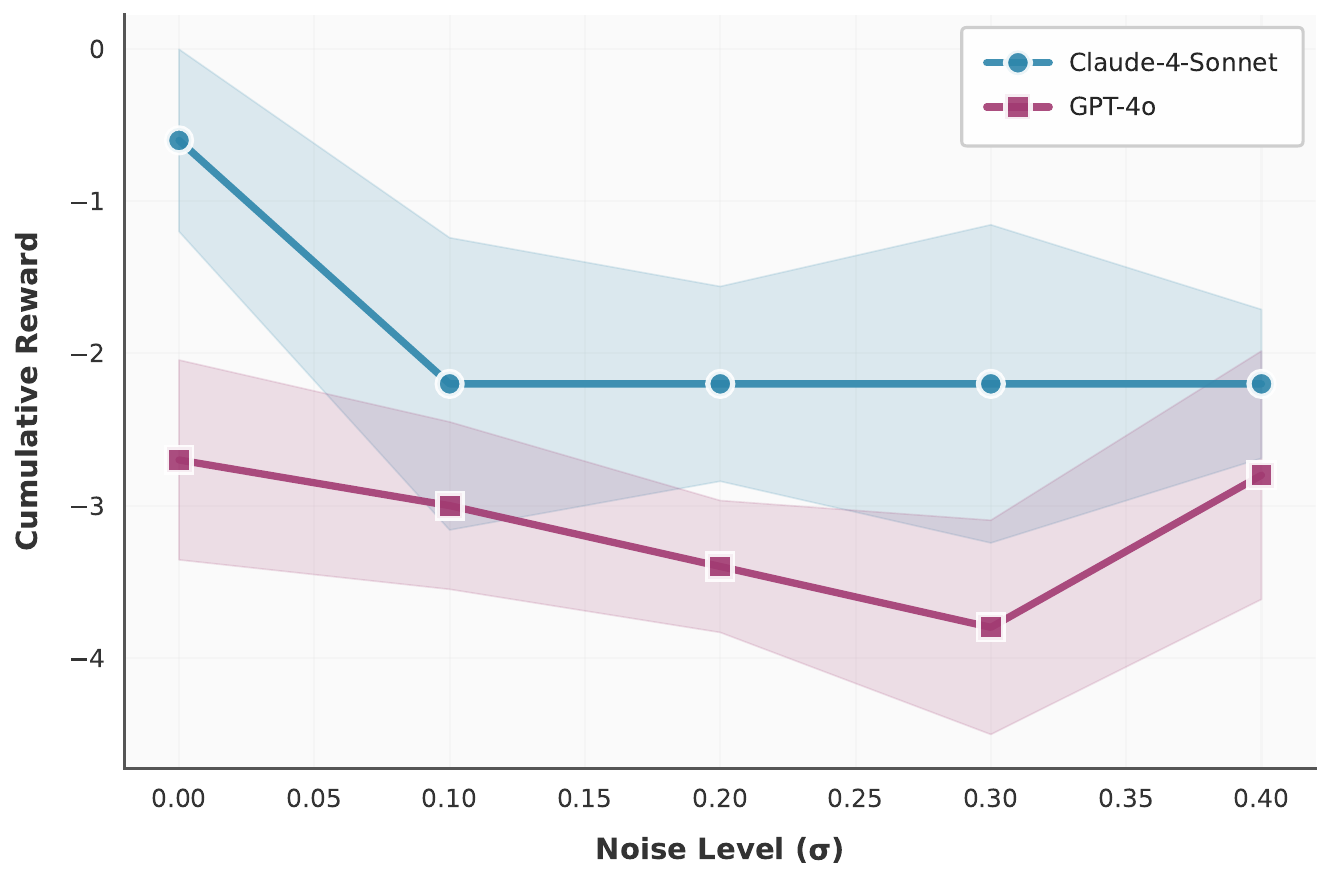}
\caption{Noise robustness analysis for Pong. Performance shows rapid degradation similar to Breakout.}
\label{fig:noise_pong}
\end{figure}

\begin{figure}[!htb]
\centering
\includegraphics[width=0.9\linewidth]{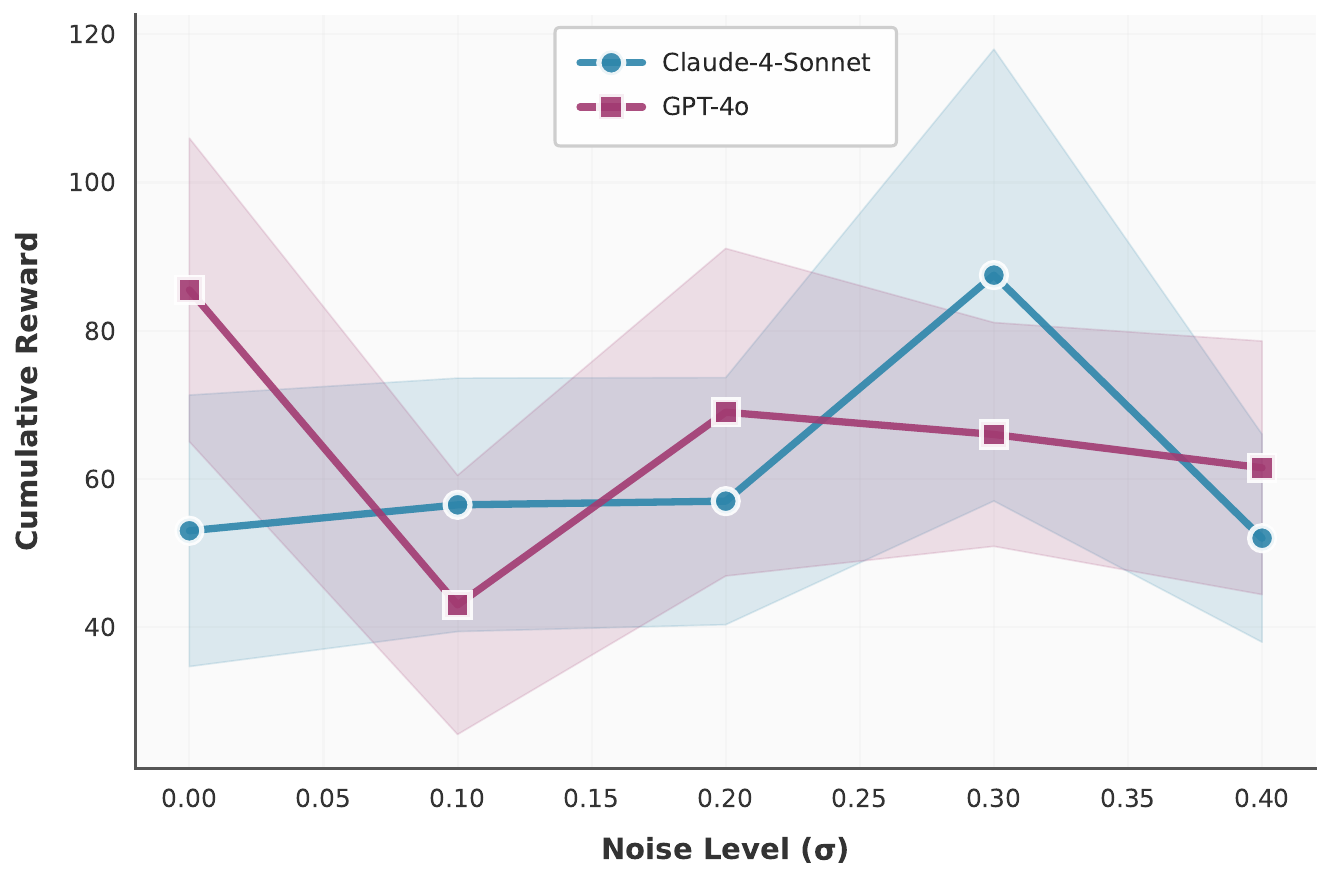}
\caption{Noise robustness analysis for Space Invaders. High variance due to dense alien formations where random firing can still score points.}
\label{fig:noise_spaceinvaders}
\end{figure}

Figure~\ref{fig:noise_pong} and Figure~\ref{fig:noise_spaceinvaders} show the noise robustness analysis for Pong and Space Invaders, complementing the Breakout results in the main paper.

\subsection{Prompts}
This section provides the complete prompts used in our experiments, including the symbolic information extraction prompt and the prompts for each pipeline.

\subsubsection{Symbolic Extraction by VLM Prompt}
This prompt is used by the Frame + Self-Extracted Symbols pipeline to extract object-centric symbolic information from the visual frame.
\begin{lstlisting}
You are an expert game frame analyzer for the game Breakout.

Your task is to detect ALL visible objects in the image
with high precision. Detect all distinct, visible objects
(like players, enemies, projectiles, items, scores). For
each object, provide its label, a tight bounding box
[x1, y1, x2, y2], and a confidence score.

Return ONLY valid JSON in the following format:
{
    "objects": [
        {
            "id": "unique_id",
            "label": "object_type_or_description",
            "coordinates": [x1, y1, x2, y2],
            "confidence": 0.95,
            "description": "brief description of the object"
        }
    ],
    "image_info": {
        "total_objects": 0,
        "frame_analysis": "brief description of what you
                           see in the frame"
    }
}
\end{lstlisting}

\subsubsection{Frame-only Pipeline Prompt}
\begin{lstlisting}
You are an expert Breakout player controlling the
ORANGE PADDLE at the bottom.

IMPORTANT: You are controlling the ORANGE PADDLE at
the bottom.

Game controls:
- Action 0: NOOP (do nothing)
- Action 1: FIRE (primary action - often shoot/serve/activate)
- Action 2: RIGHT (move right or right action)
- Action 3: LEFT (move left or left action)
- Action 4: RIGHTFIRE (combination of right + fire)
- Action 5: LEFTFIRE (combination of left + fire)

As an expert Breakout player controlling the ORANGE PADDLE
at the bottom, analyze the visual scene and choose the
optimal action.

Think step by step:
1. Observe the current state of the game
2. Predict the trajectory or movement patterns
3. Consider your strategic options
4. Choose the optimal action

Return ONLY JSON:
{
    "reasoning": "your expert analysis and decision rationale",
    "action": integer_action_code
}
\end{lstlisting}

\subsubsection{Frame + Self-Extracted Symbols Pipeline Prompt}
\begin{lstlisting}
You are an expert game player analyzing a game frame.

Game controls:
- Action 0: NOOP (do nothing)
- Action 1: FIRE (primary action - often shoot/serve/activate)
- Action 2: RIGHT (move right or right action)
- Action 3: LEFT (move left or left action)
- Action 4: RIGHTFIRE (combination of right + fire)
- Action 5: LEFTFIRE (combination of left + fire)

Current frame analysis:
- Total objects detected: 7

Detected objects with coordinates and positions:
- Object 'score_display': positioned at coordinates
  x=337, y=32, size 95x35
- Object 'score_display': positioned at coordinates
  x=462, y=32, size 95x35
- Object 'level_display': positioned at coordinates
  x=1105, y=32, size 40x35
- Object 'brick_wall': positioned at coordinates
  x=640, y=260, size 1140x120
- Object 'player_paddle': positioned at coordinates
  x=857, y=657, size 135x15

IMPORTANT: Use the symbolic information when available
and reliable, but prioritize visual reasoning if objects
are missing or the symbolic data seems incomplete.

As an expert player, analyze the scene and choose the
optimal action.

Think step by step:
1. Observe the current state of the game
2. Predict the trajectory or movement patterns
3. Consider your strategic options
4. Choose the optimal action

Return ONLY JSON:
{
    "reasoning": "your expert analysis with positional awareness",
    "action": integer_action_code
}
\end{lstlisting}

\subsubsection{Frame + Ground-Truth Symbols Pipeline Prompt}
\begin{lstlisting}
You are an expert Breakout player.

Game controls:
- Action 0: NOOP
- Action 1: FIRE
- Action 2: RIGHT
- Action 3: LEFT

Current game state (OCAtari ground truth coordinates):
Total objects: 110

Detected objects:
- Player: x=792, y=648, size=128x13
- Block: x=64, y=195, size=1152x20
- Block: x=64, y=216, size=1152x20
- Block: x=64, y=236, size=1152x20
  ... [many Object entries abbreviated]

Analyze the game state and choose the optimal action.

Think step by step:
1. Observe the current state of the game
2. Predict the trajectory or movement patterns
3. Consider your strategic options
4. Choose the optimal action

Return ONLY JSON:
{
    "reasoning": "your analysis",
    "action": integer_action_code
}
\end{lstlisting}

\subsubsection{Symbol-Only Pipeline Prompt}
\begin{lstlisting}
You are playing BREAKOUT using ONLY coordinate information
(no visual display).

Rules:
- Hit bricks with the ball to destroy them and earn points
- Don't let the ball fall below the paddle
- Y-axis: 0 (top) to 210 (bottom)
- X-axis: 0 (left) to 160 (right)
- Paddle is at bottom (y ~= 190)

STRATEGY:
- Calculate horizontal distance between ball and paddle
- Predict where ball will land when it falls (y ~= 190)
- If ball is above paddle and falling (dy > 0): position
  paddle to intercept
- If ball_x > paddle_x: move RIGHT to align
- If ball_x < paddle_x: move LEFT to align
- Account for ball velocity to predict landing position

AVAILABLE ACTIONS:
0: NOOP, 1: FIRE, 2: RIGHT, 3: LEFT

CURRENT STATE:
=== GAME STATE (Symbolic Coordinates) ===

YOUR CONTROL (Player/Paddle):
  - Position: x=47, y=189
  - Size: 16x4
BALL:
  - Position: x=53, y=147
OTHER OBJECTS (6 total):
  - BlockRow: x=8, y=57
  - BlockRow: x=8, y=63
  - BlockRow: x=8, y=69
  ... [many Object entries abbreviated]

INSTRUCTIONS:
You must make a decision based ONLY on the coordinate
data above. No visual frame is provided.

Perform spatial analysis:
1. Parse all object positions, sizes, and velocities
2. Calculate critical distances (ball-to-paddle, etc.)
3. Predict future positions using velocities
4. Identify immediate threats (enemies close, etc.)
5. Select the action that best addresses the most
   critical situation

Think step-by-step, then output:
REASONING: [Your spatial analysis and calculations]
ACTION: [number only]
\end{lstlisting}

\subsection{VizDoom Prompts}

\subsubsection{VizDoom Frame-only Pipeline Prompt}
\begin{lstlisting}
You are an expert VizDoom player in the defend_the_center
scenario.

OBJECTIVE: Shoot enemies approaching from all directions
while staying alive. You are stationary in the center.

Game controls:
- Action 0: NOOP (do nothing)
- Action 1: ATTACK (shoot)
- Action 2: TURN_LEFT (rotate left)
- Action 3: TURN_RIGHT (rotate right)

STRATEGY:
- Scan for enemies by turning left/right
- When an enemy is centered in view, shoot immediately
- Prioritize closer enemies (they appear larger)
- Keep rotating to check all directions

Analyze the visual scene and choose the optimal action.

Return ONLY JSON:
{
    "reasoning": "your analysis of enemy positions",
    "action": integer_action_code
}
\end{lstlisting}

\subsection{AI2-THOR Prompts}

\subsubsection{AI2-THOR Frame-only Pipeline Prompt}
\begin{lstlisting}
You are an AI agent in a kitchen environment (AI2-THOR).

OBJECTIVE: Find and collect food items, then place them
on the countertop.

Available actions:
- Action 0: MoveAhead (move forward)
- Action 1: MoveBack (move backward)
- Action 2: RotateLeft (turn left 90 degrees)
- Action 3: RotateRight (turn right 90 degrees)
- Action 4: LookUp (tilt camera up)
- Action 5: LookDown (tilt camera down)
- Action 6: PickupObject (pick up nearby object)
- Action 7: PutObject (place held object)

Analyze the visual scene:
1. Identify visible objects (food items, furniture)
2. Determine spatial layout of the kitchen
3. Plan navigation toward target objects
4. Execute appropriate action

Return ONLY JSON:
{
    "reasoning": "scene analysis and navigation plan",
    "action": integer_action_code
}
\end{lstlisting}

\end{document}